\pdfoutput=1

\documentclass[11pt]{article}

\usepackage[]{EACL2023}

\usepackage{times}
\usepackage{latexsym}

\usepackage[T1]{fontenc}

\usepackage[utf8]{inputenc}

\usepackage{microtype}

\usepackage{inconsolata}

\usepackage{booktabs}
\usepackage{graphicx}
\usepackage{natbib}
\usepackage{hyperref}
\usepackage{todonotes}

\title{RobBERT-2022: Updating a Dutch Language Model\\ to Account for Evolving Language Use}
\author{Pieter Delobelle \and
Thomas Winters \and
Bettina Berendt}

\author{Pieter Delobelle$^{1}$ \and Thomas Winters$^1$ \and Bettina Berendt$^{1,2,3}$\\
$^1$ Department of Computer Science, KU Leuven; Leuven.AI \\
$^2$ Faculty of Electrical Engineering and Computer Science, TU Berlin \\
$^3$ Weizenbaum Institute, Berlin\\
\texttt{firstname.lastname@kuleuven.be}}

\begin{document}

\maketitle              %
\begin{abstract}
Large transformer-based language models, e.g. BERT and GPT-3, outperform previous architectures on most natural language processing tasks. Such language models are first pre-trained on gigantic corpora of text and later used as base-model for finetuning on a particular task. Since the pre-training step is usually not repeated, base models are not up-to-date with the latest information. In this paper, we update RobBERT, a RoBERTa-based state-of-the-art Dutch language model, which was trained in 2019. First, the tokenizer of RobBERT is updated to include new high-frequent tokens present in the latest Dutch OSCAR corpus, e.g. corona-related words. Then we further pre-train the RobBERT model using this dataset. To evaluate if our new model is a plug-in replacement for RobBERT, we introduce two additional criteria based on concept drift of existing tokens and alignment for novel tokens.We found that for certain language tasks this update results in a significant performance increase. These results highlight the benefit of continually updating a language model to account for evolving language use.
\end{abstract}

\section{Introduction}

Large pre-trained transformer-based languge models have become the standard in recent years \cite{devlin2019bert,brown2020gpt,liu2019roberta}.
BERT-like models often achieve state-of-the-art results for classification and regression tasks, both on document-level as well as token-level \cite{devlin2019bert,liu2019roberta}.
Multilingual BERT models are BERT models trained on multiple languages at once, thus being able to generalize between languages, and useful for multilingual tasks.
However, monolingual models often outperform multilingual for monolingual tasks \cite{martin2019camembert,delobelle2020robbert}.

BERT-like models are usually pre-trained once, and then finetuned for many different tasks for the years to come.
However, since language evolves and the meaning of concepts begin to drift \cite{wang2010conceptdrift}, these models are continuously getting more and more outdated with respect to new concepts, trends and meanings.
To counteract this, one can pre-train a BERT model further on a more recent dataset when it emerges \cite{jin2021lifelongpretraining}.
This is similar to how BERT models are already often adapted to new domains by pre-training on large unlabeled corpora in the new domain before being finetuned to increase their performance on downstream tasks within that domain \cite{gururangan2020dontstoppretraining}.
The new domain could of course also just be a more recent dataset to further pre-train the large language model.

In this work, we use the Dutch portion of the 2022 version of the OSCAR corpus \cite{abadji2022oscar} to further pre-train the state-of-the-art Dutch BERT model RobBERT \cite{delobelle2020robbert}, which was trained on the 2019 OSCAR corpus \cite{ortizsuarez2019oscar}.
We first extend the tokenizer to include new common (sub)words, and then further pre-train RobBERT v2 to capture the recent evolution of the Dutch language.

\section{Background and Related Work}

\subsection{BERT Language Model}

Large pre-trained transformer-based language models have dominated natural language processing leaderboards for the last couple of years.
BERT-based models (e.g. BERT, BART and RoBERTa) perform exceptionally well for classification and regression tasks, such as sentiment analysis, part-of-speech detection, coreference-resolution, natural language inference, question-answering, etc \cite{rogers2020bert,liu2019roberta}.
Generally speaking, these models are only pre-trained once using a large unlabeled training dataset, and then finetuned on different downstream tasks using smaller datasets.
During the pre-training phase, the main task of the model is a form of the Cloze task called masked-language modeling (MLM).
For RoBERTa-based models, this is even the only pre-training task, since the original BERT next sentence prediction task was found to not contribute much to the training \cite{liu2019roberta}.
In the MLM task, the model has to predict the right tokens for the masked and swapped tokens of texts from the original pre-training corpus.
Using this task, the model can learn all sorts of linguistic features that are useful for downstream tasks.

\label{sec:tokenizers}
An important part of the design of a BERT-based model is the tokenizer.
The tokenizer uses the calculated vocabulary to map words to integers, which can then be used as input for the language model.
The original BERT model used a top-down WordPiece tokenizer, which segments words into subwords.
When constructing a WordPiece tokenizer, tokens are decided by combining word units in a way that increase the likelihood of the training data the most.
RoBERTa on the other hand uses byte-pair encodings (BPE), a similar method that chooses the most frequent symbol pairs \cite{sennrich2016bpe,wang2020bbpe,liu2019roberta}.
BPE and WordPiece tokenization only differs in the way symbol pairs are added to the vocabulary.

BPE tokenizers are built by looking for the most common substrings to optimally compress the dataset into as few tokens as possible.
Starting from character tokens, they continuously merge existing tokens if their combination is frequent enough, regardless of whether it makes morphological sense.
This way, larger text can on average be encoded and passed to the model.
Note that while BPE tokenizers are the most popular type of tokenizer for current language models, they have the surprising artifact of adding words multiple times, namely with and without word boundary separator (Ġ).
Different capitalizations of the first letter also cause duplicate tokens and tokens that seemingly miss their first letter. %

While multilingual models generally perform quite well, monolingual models tend to outperform them for monolingual language tasks of that particular language \cite{martin2019camembert,delobelle2020robbert}.
For Dutch, the state-of-the-art model is the RobBERT model, which was released in the first month of 2020 \citep{delobelle2020robbert}.
The model was trained on scraped internet data from the Dutch OSCAR corpus \cite{ortizsuarez2019oscar}.
Parallel works that were released only a couple of weeks before the RobBERT paper was released, are the BERTje and BERT-NL models.
They are based on the standard BERT architecture and trained on a more formal but smaller dataset \cite{devries2019bertje,brandsen2019bert}.
Thanks to the larger dataset and being based on the optimized RoBERTa architecture, the RobBERT model outperforms the other Dutch BERT models on most tasks.

\subsection{Domain Adaptation}

Further pre-training a large language model can help the model better understand the new domain of a dataset better if a large unlabeled corpus from within the domain is available.
Large language models, like BERT models, that are domain-adapted in this way increases their performance on downstream tasks.
When adapting the domain, the vocabulary of the tokenizer of the model gets extended to account for new common (sub-)words of the domain \cite{yao2021adapt,rottger2021temporaladaptation}
Adding tokens has been shown to help reliable increase the performance \cite{gu2021domain} and adapting the domain by pre-training in multiple phases %
also generally offers large performance increases \cite{gururangan2020dontstoppretraining}.
For Dutch, researchers successfully adapted the RobBERT model to the medicine~\cite{verkijk2021medicalrobbert} and law~\cite{boer2022lawrobbert} domains.

\citet{rottger2021temporaladaptation} experimented with adapting language models to more recent datasets.
They found that language models performed better on test sets of the past than the future, showing the importance of adapting to newer language use.
\citet{jin2021lifelongpretraining} also showed that further pre-training language models on more recent emerging datasets is effective for some downstream tasks.
However, it is hard to predict if an updated language model that was further pre-trained has improved or degraded performance on downstream tasks.
For example, the language model might suffer a form of catastrophic forgetting of some crucial information for particular downstream tasks, leading to unexpected degraded performance \cite{xu2020catastrophicforgetting,delange2022continual}.

\section{Method}
We propose a new RobBERT model to address the need for a more up-to-date Dutch model that takes into account many of the changes that have occurred in language usage over the past three years, including the COVID-19 pandemic. 
New terms and meanings for words have been introduced, and there is new world knowledge that RobBERT is not aware of. By retraining the model on a more recent dataset, we can ensure that it is better equipped to handle these changes.
This section will discuss new training data (\autoref{ss:data}), propose a method to extend a tokenizer's vocabulary  (\autoref{ss:vocab}) and outline the pre-training regime (\autoref{ss:pretraining}).

\subsection{Pre-training Data}\label{ss:data}
The Dutch section of the OSCAR corpus~\citep{ortizsuarez2019oscar} is based on automatically language-classified Common Crawl documents.\footnote{\url{https://oscar-corpus.com/}}
The corpus has a release with data up until January 2022, which in contrast with previous releases is now a full documents corpus instead of the previous pre-shuffled lines.
This allows models to learn longer-range dependencies, which was shown to be beneficial for some tasks~\citep{delobelle2021robbertje}.

Similar to the RoBERTa document pre-processing \cite{liu2019roberta}, we pre-processed the documents by splitting the documents into sentences and then maximizing the number of complete sentences that fit within the 512-token input limit.
Compared to simply splitting the document into 512 token chunks, this pre-processing avoids ending and starting inputs on arbitrary tokens.
This results in a training dataset of 77GB with 122,364,485 separate documents.

\subsection{Extending BPE Vocabulary}\label{ss:vocab}
Before the pre-training, we extended the vocabulary used by RobBERT's tokenizer to reflect the evolution of Dutch in the last three years.
For example, 
words like ``corona'' or ``COVID-19'' were not in the original tokenizer since they were barely used before the pandemic.
Encoding such words allows the tokenizer to more compactly represent such words within the input size of the model.

\begin{figure}
    \centering
    \includegraphics[width=\linewidth]{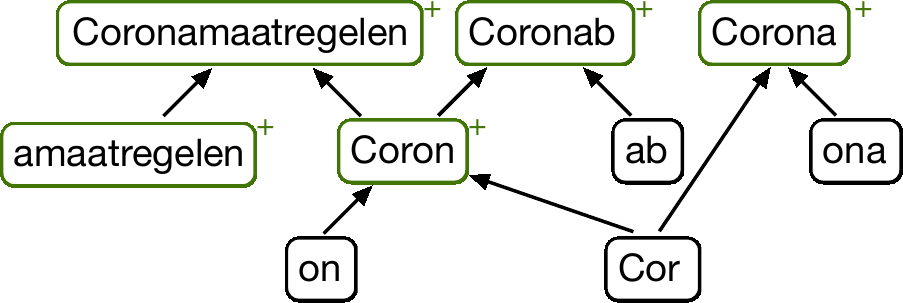}
    \caption{Illustration of how tokens related to the COVID-19 pandemic are added based on existing tokens. Notice the compound word with `maatregelen' (\emph{measures}) is not following Dutch morphological splitting rules due to how BPE token calculation works.}
    \label{fig:token_merges}
\end{figure}

To add these tokens to the vocabulary of the original tokenizer $T_O$, we first created a new BPE tokenizer $T_N$ using the process described in section \ref{sec:tokenizers} \cite{sennrich2016bpe,liu2019roberta}.
After creating the new tokenizer $T_N$, we add the tokens that are newly added compared to the previous tokenizer $T_O$ to create the merged tokenizer $T_M$.
BPE tokenizers also use a dependency graph that specifies how tokens are merged into large tokens, which also needs to be updated.
For example, Figure \ref{fig:token_merges} shows how ``Coron'' is composed of ``Cor'' and ``on'', and is used in larger tokens. %
By iterating over all the merges of $T_N$, we perform the merger if the token is in the set difference between $T_O$ and $T_N$.
All other merges are already covered by $T_O$.
This results in a new vocabulary size of 42774 for $T_M$, meaning we added 2774 tokens to $T_O$.

Interestingly, these new tokens also give insight into what (sub-)words society started using more, with some categorized examples in \autoref{tab:new_tokens_sample}.
Some of the new tokens are just added because they now cross the threshold, others just because these are now much more common words, e.g. COVID-19-related words and new brands etc.

\begin{table*}
\caption{Results on the benchmarks tasks from \citet{delobelle2020robbert} as well as the pseudo-perplexity~\citep[PPPL,][]{salazar2019mlmscoring} on the latest Dutch OSCAR corpus. Indicated results are from their respective papers.}\label{table:results}
\resizebox{\textwidth}{!}{%
\begin{tabular}{@{}lrrlllllr@{}}
\toprule
              & &                                        & \multicolumn{1}{l}{\textbf{}}       & \multicolumn{3}{c}{\textsc {\textbf{Benchmark scores}}}                                                                                        \\ \cmidrule(lr){4-8}
\textbf{Model}&  \multicolumn{1}{c}{\textbf{Params}} &\multicolumn{1}{c}{\textbf{Year}} & \multicolumn{1}{c}{\textbf{SA}} & \multicolumn{1}{c}{\textbf{CR}} & \multicolumn{1}{c}{\textbf{NER}} & \multicolumn{1}{c}{\textbf{POS}}  & \multicolumn{1}{c}{\textbf{NLI}}& \textbf{PPPL}\\ \midrule
RobBERT (v2)          & 116 M   & 2019        & $ 94.4 \pm 1.0^{*}$                   & $\mathbf{99.2} \pm 0.03^{*}$             & $\mathbf{89.1}^{*}$             & $\mathbf{96.4}\pm 0.4^{*}$ & $84.2 \pm 1.0$                            & 35.9   \\ 
BERTje           &  109~M       & 2019       & 93.0$^{***}$                  &    98.3$^{*}$          & 88.3$^{*}$            & 96.3$^{*}$ & 83.94$^{**}$                          & 144.2   \\ %

RobBERTje (Non-shuffled)         & 74 M      & 2021         & $90.2 \pm 1.2$& ${98.4} \pm 0.1$                    &  ${82.9}$            & $95.5 \pm 0.4$ & ${83.4} \pm 1.0$   & 92.1 \\
RobBERT-2022                  & 119M       & 2022         & $ \mathbf{95.1} \pm 0.9$& $97.8 \pm 0.1$                    & $87.0$             & $96.1 \pm 0.3$ &  $ \mathbf{84.9} \pm 1.0$                     & $\mathbf{8.20}$           \\ %

  \bottomrule

\end{tabular}%
}
\end{table*}

\subsection{Continuing Pre-training}\label{ss:pretraining}

To make language models account for evolved language usage, one could either pre-train from scratch, continue the pre-training on a more recent dataset or distill the model using a more recent dataset \cite{yao2021adapt}.
While RobBERT has been successfully distilled already \cite{delobelle2021robbertje}, it is not trivial to use such a distillation process in this case due to the enlarged vocabulary.
Since the new corpus is a superset of the previously used 2019 Dutch OSCAR and still includes most scraped data of the older version, we take the pre-trained weights from the original RobBERT model and continue training on the expanded corpus.
To account for the newly introduced tokens, we increased the embeddings matrix with 2774 tokens.

The pre-training was performed using the same training regime as \citet{delobelle2020robbert}, which is equivalent to RoBERTa's regime~\citep{liu2019roberta}.
We use gradient accumulation over 128 batches for an effective batch size of 1024 with one 3080 Ti, we use ADAM~\citep{kingma2014adam} as optimizer with $lr=10^{-6}$ and warmup of 1k batches.
We terminate our pre-training with early stopping based on the validation set performance, which we test every 100k steps using an MLM-specific version of perplexity~\citep[\textbf{PPPL},][]{salazar2019mlmscoring}.

\begin{table}[]
\caption{Results on VaccinChat~\citep{buhmann-etal-2022-domain}, a task on FAQ about COVID-19 vaccines. Results without $F_1$ score are reported by \citet{buhmann-etal-2022-domain}.}\label{table:results-covid}
\centering
\resizebox{0.8\linewidth}{!}{%
\begin{tabular}{@{}rll@{}}
\toprule
\multicolumn{1}{l}{\textbf{Model}}                  & \textbf{ACC}  & $\mathbf{F_1}$    \\ \midrule
\multicolumn{1}{l}{\textbf{Domain-adapted models}}  &                 &                 \\
BERTje+                                             & 77.7\%          & ---             \\
CoNTACT+                                            & \textbf{77.9\%} & ---             \\
\multicolumn{1}{l}{\textbf{General-purpose models}} &                 &                 \\
BERTje                                              & 74.7\%          & ---             \\
RobBERT v2                                          & 74.9\%          & 77.2\%          \\
RobBERT-2022                                        & \textbf{76.3\%} & \textbf{79.3\%} \\ \bottomrule
\end{tabular}
}
\end{table}

\section{Evaluation and Results}

We first evaluated the model on a variety of tasks to verify that minimal concept drift and catastrophic forgetting happened (\autoref{ss:old-eval}).
We then also evaluated its performance on two recent COVID-related tasks and performed an embedding analysis (\autoref{ss:new-eval}).

\subsection{No or minimal shift for existing concepts}\label{ss:old-eval}

We first performed an evaluation on the same RobBERT benchmark tasks as the original model to analyze the effect on existing concepts and to ensure there was minimal concept drift and no catastrophic forgetting.
A detailed description of the benchmarks can be found in \citet{delobelle2021robbertje} and our training setup is in \autoref{app:setup}.
The results of the comparisons in \autoref{table:results} show that there is no decrease in accuracy or utility on most benchmarks and even some improvement for two tasks.

Second, we test if the pseudo-perplexity on the original corpus from 2019 increased over pseudo-perplexity of the original RobBERT model, since this would indicate that the original dataset is not modelled anymore by RobBERT-2022.
The pseudo-perplexity of 9.40 is slightly higher compared to 7.76 for RobBERT, which can be an indication that our assumption—that the new dataset is a superset of the original one—is not entirely correct.
However, it is still reasonably low and shows that the new RobBERT-2022 model is still able to model the original data distribution.

\subsection{Meaningful embeddings for new concepts}\label{ss:new-eval}

We evaluated the model's embeddings for new concepts by first testing its performance on a task to classify tweets about COVID-19 measures in Belgium~\citep{scott2021covid}. %
While both the original RobBERT \cite{delobelle2020robbert} and multilingual BERT \cite{devlin2019bert} both achieve 73\% test set accuracy on this task, RobBERT-2022 achieves 75\% accuracy.
This indicates that having these new tokens with meaningful embeddings helped RobBERT-2022 better classify this recent dataset.

Secondly, we compared the predictive performance of RobBERT-2022 on 
VaccinChat~\citep{buhmann-etal-2022-domain} to 
the original RobBERT model, BERTje~\citep{devries2019bertje}, another Dutch model with data from 2019, and CoNTACT~\citep{Lemmens2022CoNTACT}, a domain-specific model for Dutch COVID-19 tweets based on RobBERT.
The results in \autoref{table:results-covid} demonstrate that the updated model benefits from the COVID-19-related tokens and training data, with an $F_1$ score of 79.3\% for the new model, compared to 77.2\% for the original model.
However, the performance of domain-adapted models BERTje+ and CoNTACT+~\citep{buhmann-etal-2022-domain} highlight that RobBERT-2022 is still a general-purpose language model and further domain-adaptation is beneficial.

\begin{figure}[t]
    \centering
    \includegraphics[width=\linewidth]{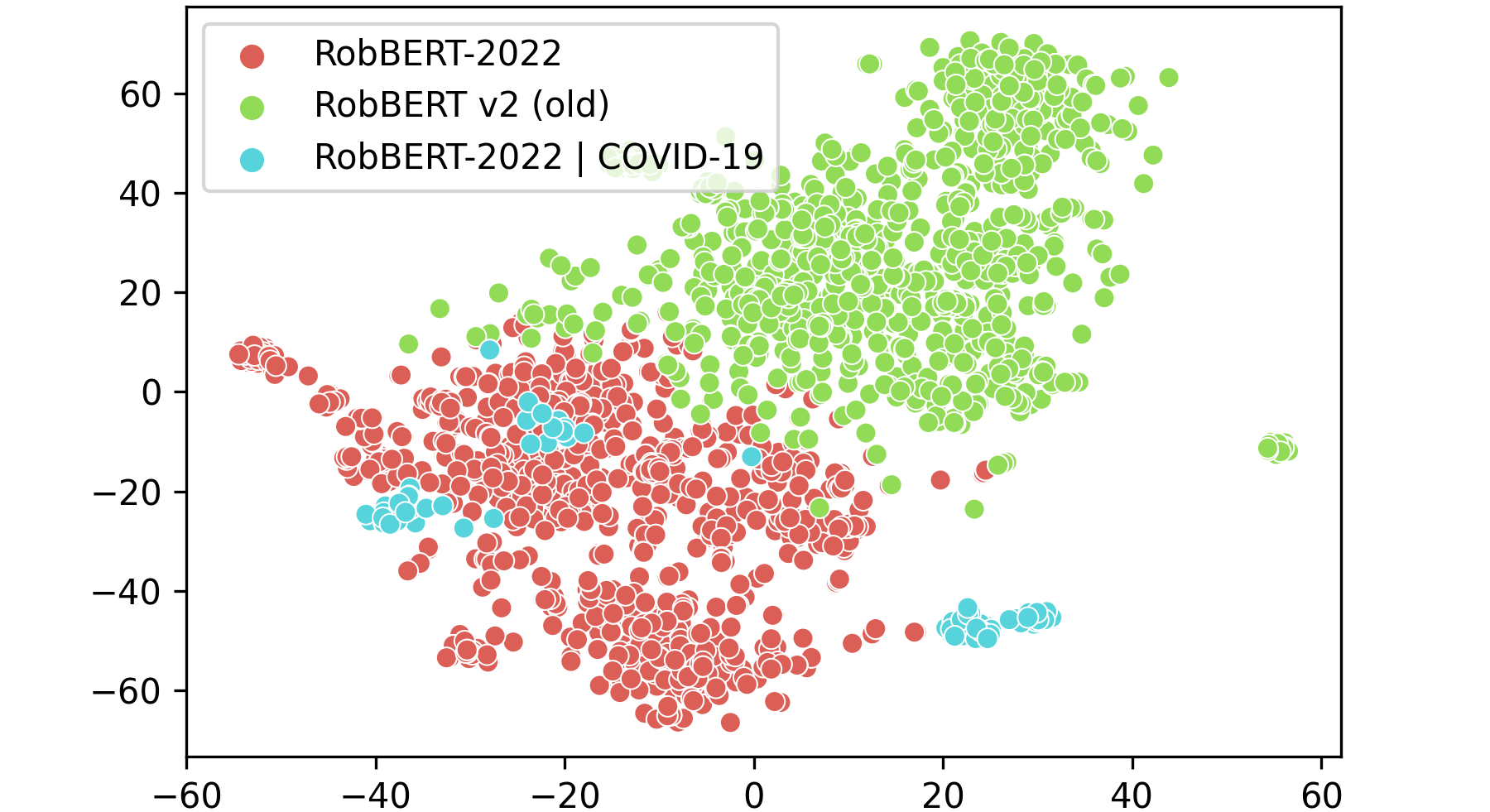}
    \caption{t-SNE visualisation of sentence embeddings for unseen sentences from the COVID-19 and OSCAR 2022 datasets. Notice the shift between both models.}
    \label{fig:embedding}
\end{figure}

Thirdly,
we evaluated the model's embeddings by performing an embeddings analysis
to test if they can capture the same relevant information 
using sentences of the validation set
(\autoref{fig:embedding}).
Interestingly, this indicates that RobBERT-2022's sentence embeddings drifted from the original.

\section{Model} %
RobBERT-2022 is available on HuggingFace as {\tt DTAI-KULeuven/robbert-2022-dutch-base}\footnote{\url{https://huggingface.co/DTAI-KULeuven/robbert-2022-dutch-base}}. %

\section{Conclusion \& Future Work}

We created an updated version of the state-of-the-art Dutch BERT-like model RobBERT by extending its vocabulary and further pre-training the model.
Our analysis shows that further pre-training on a recent dataset increases performance on more recent tasks while maintaining performance on older tasks.
This indicates that updating the model in the future on more recent training corpora is a fruitful endeavor.
Since RobBERT was already successfully distilled in the RobBERTje model~\cite{delobelle2021robbertje}, it would be interesting evaluate the performance difference of a distilled RobBERT-2022 model with a further pre-trained RobBERTje model.
We hope that like its predecessor, this new RobBERT-2022 model helps to provide the Dutch NLP community with a model that can be used on datasets relying on more recent information.

\bibliography{anthology,custom}
\bibliographystyle{acl_natbib}

\clearpage
\appendix

\begin{table*}[]
\centering
\resizebox{\textwidth}{!}{
\begin{tabular}{llllllllll}
\toprule
 & \textbf{User Interface}
 & \textbf{Metadata}
 & \textbf{ Normal}
 & \textbf{Locations}
 & \textbf{COVID-19}
 & \textbf{Names}
 & \textbf{Brands}
 & \textbf{GDPR / cookie}
 & \textbf{Time} \\ \hline
 &
 \begin{tabular}[t]{l@{}}
 Aanmelden\\ About\\ Admin\\ Advertentie\\ Antwoorden\\ Controleer\\ CONTACT\\ FAQ\\ Filter\\ Follow\\ Geschiedenis\\ Inloggen\\ Instellingen\\ Kalender\\ Leden\\ Loading\\ Nieuwsbrief\\ Quote\\ Reageren\\ Search\\ Verstuur\\ voorkeuren \\ ĠComment \\ Ġinbox \\ ĠReply \\ ĠSorteer \\ ĠToevoegen 
 \end{tabular}
 &
 \begin{tabular}[t]{l@{}}
 Beoordeling \\ Bestand \\ Breedte \\ Datum\\ Genre \\ Gerelateerd \\ Geslacht \\ Gewicht \\ Ingredi\"enten \\ ISBN \\ language \\ Locatie\\ Maand \\ Onderwerp \\ Openingstijden \\ Related \\ Resultaten \\ Telefoonnummer \\ Titel \\ Trefwoorden \\ uplicatiedatum \\ VANAF \\ Voornaam \\ Woonplaats \\ ĠCount \\ ĠPublicatiedatum \\ ĠUitspraakdatum
 \end{tabular}
 &
 \begin{tabular}[t]{l@{}} 
 Abonnement \\ Aluminium \\ Anoniem \\ begraafplaats \\ Belasting \\ Campus \\ Communicatie \\ Cultuur \\ Dienst \\ Eten \\ Fietsen \\ forums \\ gieter \\ Kantoor \\ missies \\ Paragraaf \\ Podcast \\ Prins \\ Samenvatting \\ Tips \\ Vacatures \\Wetenschap \\ woordenboek \\ ĠBruiloft \\ ĠPastoor \\ Ġraamdecoratie \\ Ġsurfgedrag
 \end{tabular}
 & 
 \begin{tabular}[t]{l@{}} 
 Amersfoort \\ Belgium \\ Bergen \\ Breda \\ Eindhoven \\ Finland \\ huisstraat \\ Maastricht \\ Nijmegen \\ Postbus \\ Provincie \\ Wijk \\ yprus \\ Zwolle \\ ĠAppeldoorn \\ ĠDorpstraat \\ ĠHoofdweg \\ ĠKerkhof \\ ĠKerkhoflaan \\ ĠLambertuskerk \\ ĠLichtenvoorde \\ ĠMolenweg \\ ĠNieuwkerk \\ ĠRepublic \\ ĠSchoolstraat \\ ĠZuydewijn
 \end{tabular}
 &    
 \begin{tabular}[t]{l@{}} 
 amaatregelen \\ andemie \\ avirus \\ Coron \\ Corona \\ COVID \\ Overleg \\ Kap \\ Lucht \\ pleegkundige \\ Wandel \\ 
 Ġcoron \\ Ġcorona \\ ĠCorona \\ Ġcoronamaatregelen \\ ĠCoronavirus \\ Ġcoronavirus \\ ĠCovid \\ Ġcovid \\ ĠCOVID \\ Ġgevaccin \\ Ġmondkapje \\ Ġmondmasker \\ Ġquarantaine \\ Ġvaccineren \\ Ġwebinar
 \end{tabular}
 &     
 \begin{tabular}[t]{l@{}} 
 Dirk \\ Gelder \\ Kim \\ Marcel \\ Robert \\ Sinterklaas \\ Thomas  \\ ĠBaudet \\ĠBrard \\ ĠDaphne \\ ĠDuncan \\ ĠFroukje \\ ĠHilton \\ ĠHuub \\ ĠJoosten \\ ĠLaurentien \\ ĠLeontine \\  ĠLieke \\ ĠMaes \\ ĠMaxima \\ ĠPatty \\ ĠRamses \\ ĠRoelvink \\ ĠSchneider \\ ĠSchuurman\\ ĠZegers
 \end{tabular}
 
 &  

 \begin{tabular}[t]{l@{}} 
 Android \\ CAPTCHA \\ Fox \\ google \\ HEMA \\ Instagram \\ iPhone \\ JavaScript \\ Linkedin \\ LinkedIn \\ NOS \\ phpBB \\ Pinterest \\ RTL \\ Toyota \\ Twitter \\ Wiki \\ wikipedia \\ WordPress \\ YouTube \\ Ġdeezer \\ ĠKnack \\ ĠPremiere \\ ĠTeams \\ ĠTikTok \\ ĠTrends \\ ĠTROS
 \end{tabular}
 &   
 \begin{tabular}[t]{l@{}}  
 Accepteren \\ Analytische \\ cookie \\ Cookiebeleid \\ cookielawinfo \\ Cookies \\ isclaimer \\ Necessary \\ necessary \\ Noodzak \\ Noodzakelijk \\ noodzakelijke \\ Noodzakelijke \\ Privacy \\ privacy \\ Privacybeleid \\ Privacyverklaring \\ vereist \\ Ġaccepteer \\ Ġconsent \\ Ġcookiebeleid \\ ĠDisclaimer \\ Ġfunctionalities \\ ĠFunctionele \\ Ġgebruikerservaring \\ Ġmandatory \\ ĠPrivacyverklaring
 \end{tabular}
 &  
 \begin{tabular}[t]{l@{}}  
 anuary \\ april \\ augustus \\ December \\ december \\ dinsdag \\ donderdag \\ ebruary \\ januari \\ juli \\ Juni \\ maart \\ november \\ November \\ oktober \\ riday \\ September \\ september \\ vrijdag \\ woensdag \\ zaterdag \\ ĠFeb \\ ĠFebruary \\ ĠJanuary \\ ĠJune \\ ĠKoninginnedag \\ ĠPrinsjesdag
 \end{tabular}
 \\
 \bottomrule
\end{tabular}
}

\caption{Some examples of new tokens added to RobBERT-2022 that were not in RobBERT v2, grouped by theme}
\label{tab:new_tokens_sample}
\end{table*}

\section{Experimental setup}
\label{app:setup}

\subsection{Sentiment Analysis (SA)}
We evaluate sentiment analysis on the Dutch Book Review Dataset~\citep{vanderburgh2019dbrd} with standard splits.
Our experiment consists of one run with the following hyperparameters:

\begin{itemize}
  \setlength\itemsep{0.3ex}
    \item Number of gpus: 1 (1080 Ti)
    \item adafactor: False
    \item adam beta1: 0.9
    \item adam beta2: 0.999
    \item adam epsilon: 1e-08
    \item deepspeed: None
    \item fp16: False
    \item gradient acc. steps: 8
    \item lr: $ 10^{-4}$
    \item lr scheduler type: LINEAR
    \item num train epochs: 10
    \item optimizer: ADAMW
    \item batch size: 4
    \item seed: 1
    
    \item warmup ratio: 0.0
    \item warmup steps: 20
    \item weight decay: 0.05
    
\end{itemize}

\subsection{Co-reference Resolution (CR)}
We run five randomized training run on the `die'-`dat' disambiguation task by \citet{allein2020diedat}, where we vary the learning rate, number of gradient accumulation steps and weight decay.
We limit our datasets for computational reasons, to 50k (training), 3k (validation) and 30k (testing).

\begin{itemize}
  \setlength\itemsep{0.3ex}
    \item Number of gpus: 1 (1080 Ti)
    \item adafactor: False
    \item adam beta1: 0.9
    \item adam beta2: 0.999
    \item adam epsilon: 1e-08
    \item deepspeed: None
    \item fp16: False
    \item gradient acc. steps: $\{2, 4, 8, 16\}$. Best: 2
    \item lr: $[10^{-6},10^{-4}]$. Best:  $5.29 \cdot 10^{-5}$
    \item lr scheduler type: LINEAR
    \item num train epochs: 1
    \item optimizer: ADAMW
    \item batch size: 8
    \item seed: 1
    
    \item warmup ratio: 0.0
    \item warmup steps: 20
    \item weight decay: $[0.0, 0.1]$. Best: 0.09
    
\end{itemize}

\subsection{Named Entity Recognition (NER)}
We evaluate NER on the CoNLL dataset with an experiment that consists of 10 runs with Bayesian optimisation (TPE) with the following hyperparameters, where we vary the learning rate, number of gradient accumulation steps and weight decay. We select the best-performing model based on the $F_1$ score on a separate validation set before testing this model the test set.

\begin{itemize}
  \setlength\itemsep{0.3ex}
    \item Number of gpus: 1 (1080 Ti)
    \item adafactor: False
    \item adam beta1: 0.9
    \item adam beta2: 0.999
    \item adam epsilon: 1e-08
    \item deepspeed: None
    \item fp16: False
    \item gradient acc. steps: $\{1, 2, 4, 8, 16, 32\}$. 
    
    Best: 4
    \item lr: $[10^{-6},10^{-4}]$. Best:  $7.89 \cdot 10^{-5}$
    \item lr scheduler type: LINEAR
    \item num train epochs: 10
    \item optimizer: ADAMW
    \item batch size: 8
    \item seed: 1
    
    \item warmup ratio: 0.0
    \item warmup steps: 20
    \item weight decay: $[0.01, 0.1]$. Best: 0.08
    
\end{itemize}

\subsection{Part-of-speech (POS) tagging}
We also perform 10 runs with Bayesian optimisation (TPE) with the following hyperparameters, where we vary the learning rate, number of gradient accumulation steps and weight decay. We select the best-performing model based on the $F_1$ score on a separate validation set before testing this model the test set. 

\begin{itemize}
  \setlength\itemsep{0.3ex}
    \item Number of gpus: 1 (1080 Ti)
    \item adafactor: False
    \item adam beta1: 0.9
    \item adam beta2: 0.999
    \item adam epsilon: 1e-08
    \item deepspeed: None
    \item fp16: False
    \item gradient acc. steps: $\{1, 2, 4, 8, 16, 32\}$. 
    
    Best: 1
    \item lr: $[10^{-6},10^{-4}]$. Best:  $4.5 \cdot 10^{-5}$
    \item lr scheduler type: LINEAR
    \item num train epochs: 10
    \item optimizer: ADAMW
    \item batch size: 8
    \item seed: 1
    
    \item warmup ratio: 0.0
    \item warmup steps: 20
    \item weight decay: $[0.01, 0.1]$. Best: 0.06
    
\end{itemize}

\subsection{Natural Language Inference (NLI)}
Our experiment consists of 5 runs with the following hyperparameters, where most are fixed and the learning rate, weight decay and the number of gradient accumulation steps are randomly selected from the specified ranges.

\begin{itemize}
  \setlength\itemsep{0.3ex}
    \item Number of gpus: 1 (1080 Ti)
    \item adafactor: False
    \item adam beta1: 0.9
    \item adam beta2: 0.999
    \item adam epsilon: 1e-08
    \item deepspeed: None
    \item fp16: False
    \item gradient acc. steps: $\{2, 4, 8, 16\}$. Best: 8
    \item lr: $[10^{-6},10^{-4}]$. Best: $4.8 \cdot 10^{-5}$
    \item lr scheduler type: LINEAR
    \item num train epochs: 10
    \item optimizer: ADAMW
    \item batch size: 8
    \item seed: 1
    
    \item warmup ratio: 0.0
    \item warmup steps: 20
    \item weight decay: $[0, 0.1]$. Best: 0.025
    
\end{itemize}

\subsection{COVID-19 topics}
We take the task of extracting the topic of a Tweet, as categorised and labeled by \citet{scott2021covid}, with the following topics as possible labels:
\begin{itemize}
  \setlength\itemsep{0.3ex}
    \item closing-horeca
    \item testing
    \item schools
    \item lockdown
    \item quarantine
    \item curfew
    \item masks
    \item vaccine
    \item other-measure
    \item not-applicable
\end{itemize}

Our experiment consists of 5 runs with the following hyperparameters, where most are fixed and the learning rate, weight decay and the number of gradient accumulation steps are randomly selected from the specified ranges.
\begin{itemize}
  \setlength\itemsep{0.3ex}
    \item Number of gpus: 1 (1080 Ti)
    \item adafactor: False
    \item adam beta1: 0.9
    \item adam beta2: 0.999
    \item adam epsilon: 1e-08
    \item deepspeed: None
    \item fp16: False
    \item gradient acc. steps: $\{2, 4, 8, 16\}$. Best: 2
    \item lr: $[10^{-6},10^{-4}]$. Best: $9.75 \cdot 10^{-5}$
    \item lr scheduler type: LINEAR
    \item num train epochs: 10
    \item optimizer: ADAMW
    \item batch size: 8
    \item seed: 1
    
    \item warmup ratio: 0.0
    \item warmup steps: 20
    \item weight decay: $[0, 0.1]$. Best: 0.065
    
\end{itemize}

\subsection{COVID-19 topics}
We follow the train, validation and test sets as introduced by \citet{buhmann-etal-2022-domain} on \url{https://huggingface.co/datasets/clips/VaccinChatNL}.
Our experiment consists of 5 runs with the following hyperparameters for both RobBERT v2 and RobBERT-2022, where most are fixed and the learning rate, weight decay and the number of gradient accumulation steps are randomly selected from the specified ranges.
\begin{itemize}
  \setlength\itemsep{0.3ex}
    \item Number of gpus: 1 (1080 Ti)
    \item adafactor: False
    \item adam beta1: 0.9
    \item adam beta2: 0.999
    \item adam epsilon: 1e-08
    \item deepspeed: None
    \item fp16: False
    \item gradient acc. steps: $\{2, 4, 8, 16\}$. \\
    Best RobBERT v2: 2\\
    Best RobBERT-2022: 4 
    \item lr: $[10^{-6},10^{-4}]$. \\
    Best RobBERT v2: $6.821 \cdot 10^{-5}$\\
    Best RobBERT-2022: $6.824 \cdot 10^{-5}$
    \item lr scheduler type: LINEAR
    \item num train epochs: 5
    \item optimizer: ADAMW
    \item batch size: 8
    \item seed: 1
    
    \item warmup ratio: 0.0
    \item warmup steps: 20
    \item weight decay: $[0, 0.1]$. \\
    Best RobBERT v2: $0.066$\\
    Best RobBERT-2022: $0.039$
    
\end{itemize}

\end{document}